\documentclass[journal]{IEEEtran}

\ifCLASSINFOpdf
\else
\fi
\usepackage{pifont}
\usepackage{bbding}
\usepackage{tikz}
\usepackage{subcaption}
\usepackage{graphicx}
\usepackage{booktabs}
\usepackage{makecell}
\usepackage{multirow}
\usepackage{pifont}
\usepackage{amsmath}
\usepackage{url}
\usepackage{mathtools}
\usepackage{amssymb}
\usepackage{amsfonts}
\usepackage{xcolor}
\usepackage{colortbl}
\usepackage{algorithm}
\usepackage{algorithmic}
\usepackage{threeparttable}
\usepackage{stfloats}
\begin{document}

\title{\LARGE \bf Towards All-Day Perception for Off-Road Driving: A Large-Scale Multispectral Dataset and Comprehensive Benchmark}

\author{Shuo Wang$^{1}$, Jilin Mei$^{1}$, Wenfei Guan$^{1}$, Shuai Wang$^{1}$, Yan Xing$^{2}$, Chen Min$^{1,\dag}$, Yu Hu$^{1,\dag}$
\thanks{$^{1}$Research Center for Intelligent Computing Systems, Institute of Computing Technology, Chinese Academy of Sciences, Beijing, 100190, China.}%
\thanks{$^{2}$Beijing Institute of Control Engineering, Beijing, 100194, China.}%
\thanks{$^{\dag}$Corresponding authors: Chen Min and Yu Hu, \{mincheng, huyu\}@ict.ac.cn}%
}

\maketitle
\thispagestyle{empty}
\pagestyle{empty}

\begin{abstract}
Off-road nighttime autonomous driving suffers from unreliable visible-light perception, making infrared modality crucial for accurate freespace detection. However, progress remains limited due to the scarcity of annotated infrared off-road datasets and the inter-frame inconsistencies inherent to current single-frame methods. To address these gaps, we present the IRON dataset, which, to our knowledge, is the first large-scale infrared dataset for off-road temporal freespace detection under all-day conditions, with strong support for nighttime perception. The dataset comprises 24,314 densely annotated infrared images with synchronized RGB images in diverse scenes and different light conditions. Building upon this dataset, we propose IRONet, a novel flow-free framework for temporal freespace detection that addresses inter-frame inconsistencies by aggregating historical context via a memory-attention mechanism and a carefully designed mask decoder. On our IRON dataset, IRONet achieves state-of-the-art performance, reaching 82.93\%(+1.19\%) IoU and 90.66\%(+0.71\%) F1 score at real-time inference. Remarkably, IRONet also exhibits robust generalization to RGB modalities on ORFD and Rellis datasets. Overall, our work establishes a foundation for reliable all-day off-road autonomous driving and future research in infrared temporal perception. The code and IRON dataset are available at \url{https://github.com/wsnbws/IRON}.
\end{abstract}

\section{Introduction}
\begin{figure}[!t]
    \centering
    \includegraphics[width=3.4in]{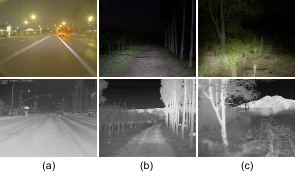}
    \caption{Comparison of RGB and IR perception under nighttime conditions. Top: RGB images; Bottom: IR images. (a) On-road nighttime scene. (b) Typical off-road nighttime scene. (c) Challenging off-road nighttime scene.}
    \label{fig:perception_contrast}
    \vspace{-5mm}
\end{figure}

Freespace detection for autonomous driving confronts a critical challenge in nighttime off-road scenes, where insufficient illumination makes visible cameras unreliable. As shown in Fig.~\ref{fig:perception_contrast}(a), abundant artificial illumination in cities ensures the reliability of visible cameras. However, the perceptual range of visible cameras is
severely limited at nighttime shown in (b). Even worse, scenarios in (c) involving turning or occlusions further exacerbate this issue. Under such conditions, the infrared modality could offer more perceptual information compared to the visible modality~\cite{shin2023deep, kim2024causal, jang2025multispectral}. Thus, more effective infrared-based freespace detection is essential for reliable off-road autonomous driving, especially under nighttime conditions.

For infrared freespace detection, an infrared dataset is indispensable, as direct model transfer from RGB to infrared would cause performance degradation~\cite{huang2025infrared, yuan2025unirgb, li2025progressive}. However, existing off-road perception research primarily focuses on RGB datasets~\cite{jiang2021rellis, min2022orfd, mortimer2024goose, sharma2022cat}, leaving infrared largely underexplored. M2P2~\cite{datar2024m2p2} recognizes the value of infrared images but lacks corresponding dense annotations. Consequently, the scarcity of annotated infrared datasets becomes the main obstacle to advancing infrared off-road freespace detection.
To address this scarcity of infrared data, we construct the IRON dataset, providing the first large-scale dataset for infrared off-road freespace detection.  The dataset consists of 35 video sequences captured across diverse off-road terrains under various light conditions. In total, our IRON dataset contains approximately 24k infrared images with pixel-wise freespace annotations and aligned timestamps. To facilitate future research on multi-modal fusion, each infrared image is also accompanied by a synchronized RGB image.

Apart from the dataset, another critical challenge is inter-frame inconsistencies (temporal flickering and spatial fragmentation)~\cite{zhuang2023video, hesham2025exploiting}. Complex off-road roads (uneven terrain, frequent turns, etc.) and color-limited infrared images aggravate this problem. Yet, current off-road freespace detection methods rely on single-frame perception~\cite{wang2021sne, li2021bifnet, li2024roadformer, sun2025rod}, resulting in poor temporal consistency. Fortunately, in other fields, modeling temporal information, whether based on optical flow~\cite{weng2023mask, fedynyak2023global} or attention mechanisms~\cite{liu2023petrv2, xu2024rgb, baek2025evolve}, has proven effective in mitigating such inconsistencies.

Inspired by this, we build a temporal framework, IRONet, upon our IRON dataset to improve inter-frame consistency and segmentation accuracy in infrared off-road freespace perception. The framework employs a pretrained Vision Transformer (ViT) backbone and a fusion module to extract multi-scale infrared features. Then, IRONet incorporates a mask-aware temporal queue and an attention mechanism to propagate contextual information from historical frames efficiently. These modules selectively aggregate freespace-relevant features, mitigating segmentation flicker and fragmentation without incurring the computational cost of explicit motion estimation. Furthermore, we introduce a novel mask decoder featuring two critical components: one that decouples mask token from task-specific semantics to strengthen memory modules supervision, and another that performs semantic compensation on mask token to guarantee continuous and coherent freespace tracking across frames. Extensive experiments demonstrate that the IRONet framework significantly outperforms existing approaches and establishes a new state-of-the-art in infrared freespace detection.

Our work achieves three main contributions:
\begin{itemize}
\item \textbf{IRON Dataset.} We introduce the first large-scale infrared dataset for temporal off-road freespace detection, covering diverse scenes and illumination conditions with dense pixel-wise annotations and synchronized RGB data.

\item \textbf{IRONet Framework.} We propose a temporal segmentation framework that aggregates multi-frame context via memory attention, achieving temporally stable and spatially coherent freespace detection.

\item \textbf{Comprehensive Benchmark and Evaluation.} IRONet achieves 82.93\% IoU at 32 FPS on IRON, and generalizes robustly to RGB modalities, establishing a strong benchmark for infrared temporal freespace detection.
\end{itemize}

\vspace{-4mm}
\section{Related Work}

\subsection{Off-Road and Infrared Datasets}

Off-road datasets differ from urban benchmarks such as Cityscapes~\cite{cordts2016cityscapes} and KITTI~\cite{geiger2013vision} by focusing on traversability in unstructured environments with irregular terrain, vegetation, and adverse weather. Early work such as RUGD~\cite{wigness2019rugd} provides RGB-based annotations, while RELLIS-3D~\cite{jiang2021rellis} extends to LiDAR, RGB, and infrared across 13,556 scans for 3D analysis. ORFD~\cite{min2022orfd} further introduces pixel-level freespace labels with synchronized LiDAR-RGB pairs. More recent datasets, including TartanDrive-V2~\cite{sivaprakasam2024tartandrive} and ORAD-3D~\cite{min2025advancing}, improve scale, temporal coverage, and annotation fidelity, while synthetic data such as OffRoadSynth~\cite{malek2024offroadsynth} complements real-world variability. Infrared datasets primarily address low-light perception. Urban benchmarks such as KAIST Multispectral~\cite{choi2018kaist} and LLVIP~\cite{jia2021llvip} enable RGB-IR fusion for pedestrian detection, with MFNet~\cite{ha2017mfnet} and FLIR ADAS~\cite{flir_adas} extending driving scenarios. M3FD~\cite{liu2022target} further targets adverse weather. In contrast, off-road IR datasets remain limited; M2P2~\cite{datar2024m2p2} provides synchronized RGB-D, Infrared, LiDAR, IMU, and GPS streams but lacks dense semantic annotations. A summary is given in Table~\ref{tab:dataset_contrast}.
\vspace{-2mm}
\subsection{Freespace Detection and Temporal Modeling} Monocular freespace detection has advanced from CNN encoder-decoder models~\cite{ronneberger2015u, chen2018encoder} to Vision Transformer architectures~\cite{xie2021segformer, cheng2022masked}, with SAM-derived frameworks~\cite{kirillov2023segment, sun2025rod} further pushing accuracy and efficiency in off-road scenarios. To compensate for monocular depth ambiguity, RGB-LiDAR fusion has been extensively explored~\cite{fan2020sne, wang2021sne, li2021bifnet, ye2023m2f2, li2024roadformer}, with recent models demonstrating that complementary modalities are critical in unstructured environments. Temporal modeling has followed a parallel trajectory in video segmentation. Optical flow-based feature warping~\cite{teed2020raft, huang2022flowformer} improves frame consistency but introduces substantial computational overhead and domain gap. This has motivated a shift toward flow-free spatio-temporal attention, exemplified by SAM2~\cite{ravi2024sam} and its extensions for long sequences and low-contrast scenarios~\cite{ding2025sam2long, cuttano2025samwise, zhang2025camosam2}. Despite these advances, temporal modeling remains absent from off-road freespace detection, where low-contrast infrared appearances and unstructured terrain make single-frame predictions particularly prone to flickering and fragmentation. Our work bridges this gap by introducing a flow-free memory-attention framework specialized for infrared off-road video segmentation.

\begin{table*}[!t]
    \caption{Comparison of Different Autonomous Driving Datasets.}
    \label{tab:dataset_contrast}
    \centering
    \begin{threeparttable}
    \renewcommand{\arraystretch}{1.5}
    \begin{tabular}{c|llccccc|ccc}    
    \hline
     & \textbf{Dataset} & \textbf{Modality} & \textbf{Size} & \textbf{Rate} & \textbf{Align} & \textbf{IR range} & \textbf{Resolution} & \textbf{Train} & \textbf{Test} & \textbf{Annotation Task} \\
    \hline 
    \multirow{5}{*}{\rotatebox[origin=c]{90}{\textbf{On-road}}}
    & SCUT~\cite{xu2019benchmarking} & I  &  211k & 25 & NA & NA & $384\times288$  & 108k & 103k & Pedestrian Detection \\
    & KAIST~\cite{choi2018kaist}  & R,I  & 95,328 & 20 & \checkmark & 7.5-13  & $640\times480$ & 50,187 & 45,141 & Pedestrian Detection\\
    & FLIR~\cite{flir_adas} & R,I  & 14,452 & NA & \ding{55} & 7.5-13.5  & $640\times512$ & 8,862 & 4,224 & Object Detection\\
    & MFNet~\cite{ha2017mfnet} & R,I  & 1,569 & NA & \checkmark & 8-14  & $640\times480$ & 785 & 393 & Multi-Class Segmentation\\
    & InfraParis~\cite{franchi2024infraparis} & R,I,D & 7,301 & NA & \checkmark & 8-14 & NA & 6,568 & 571 & Multi-task (Seg., Det.)\\
    \hline
    \multirow{6}{*}{\rotatebox[origin=c]{90}{\textbf{Off-road}}}
    & RUGD~\cite{wigness2019rugd} & R  & 7,546 & 15 & NA & NA & $1376\times1110$ & 4,759 & 1,964 & Multi-Class Segmentation \\
    & RELLIS-3D~\cite{jiang2021rellis} & R,L & 6,000 & 10 & \checkmark & NA & $1920\times1200$ & 3,302 & 1,672 & Multi-Class Segmentation \\
    & ORFD~\cite{min2022orfd} & R,L & 12,198 & NA &\checkmark & NA & $1280\times704$ & 8,398 & 2,555 & Freespace Detection \\
    & M2P2~\cite{datar2024m2p2} & R,I,L  & 36,000 & 10 & \checkmark & 8-14  & $1280\times1024$ & NA & NA & Metric Depth, Trajectory  \\
    & Goose~\cite{mortimer2024goose} & R,I,L & 10,000 & 10 & \checkmark & 0.7-1.0 & $1024\times768$ & 7,830 & 1,210 & Multi-Class Segmentation \\
    & IRON (Ours) & R,I & 24,314 & 2.5 & \checkmark & 8-14 & $640\times512$ & 20,022 & 4,292 & Temporal Freespace Detection \\
    \hline  
    \end{tabular}
    \vspace{1mm}
    \begin{tablenotes}
        \footnotesize
        \item \textbf{NOTE:} R: RGB, I: Infrared, L: LiDAR, NA: Not available or not applicable, Align: temporally and spatially synchronized across modalities, IR range: Infrared camera wavelength range ($\mu$m), Rate: Frame rate (Hz), Resolution: IR image resolution.
    \end{tablenotes}
    \end{threeparttable}
    \vspace{-2mm}
\end{table*}

\section{Dataset Construction}
To address the scarcity of temporal infrared data for off-road navigation, we introduce IRON, a dataset designed for developing and evaluating robust temporal freespace detection under real-world conditions.

\vspace{-2mm}
\subsection{Data Acquisition and Scene Diversity} 
\begin{figure}[!b]
    \centering
    \hspace{-3mm}
    \includegraphics[width=3.3in]{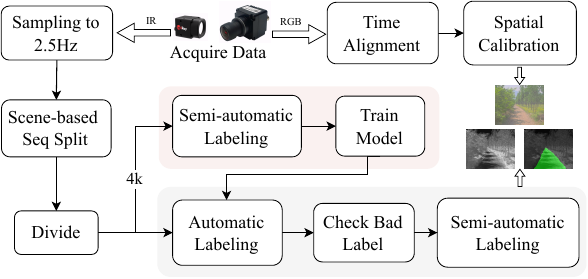}
    \caption{Data Processing and Annotation Pipeline.}
    \label{fig:data_process}
\end{figure}

\begin{figure*}[!b]
    \vspace{-3mm}
    \centering
    \includegraphics[width=7in]{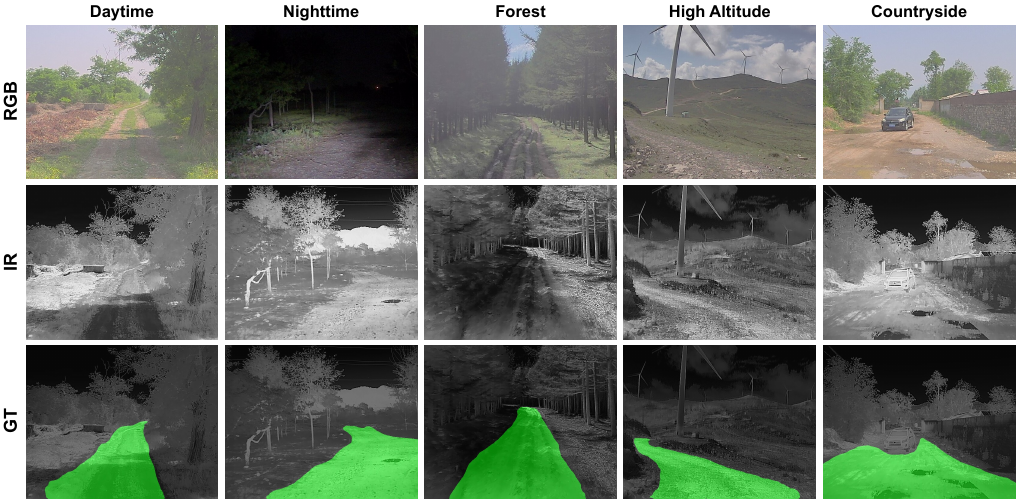}
    \caption{Samples from our IRON dataset. Each column shows a different scene, illustrating the diversity of environments and light conditions. From top to bottom: RGB images, infrared images, and freespace annotations.}
    \label{fig:dataset_overview}
\end{figure*}

The data acquisition platform comprises an infrared camera and a visible camera, rigidly mounted on a common platform at the front of the vehicle. The infrared camera (IRpilot640X~\cite{rayvision_thermal}) captures $640 \times 512$ images with a $32^\circ \times 26^\circ$ FOV in the 8--14\,$\mu$m band. The RGB images are acquired at $1920 \times 1080$ resolution by one of two cameras (SG2-AR0233C-5200-G2A-H100F1A or SG2N1BACF~\cite{seeedstudio2025}), selected according to the scene's field-of-view requirements. Temporal alignment and spatial calibration are performed to ensure cross-modal consistency, as illustrated in Fig.~\ref{fig:data_process}. Using this platform, we recorded 35 sequences at 50\,Hz across diverse off-road environments, countryside roads, forest trails, and high-altitude areas, under daytime, dusk, and nighttime illumination as shown in Fig.~\ref{fig:dataset_overview}, covering realistic conditions where infrared imaging offers a clear perceptual advantage over visible cameras.

\subsection{Dataset Annotations and Split}
Fig.~\ref{fig:data_process} illustrates the overall data processing pipeline. Initially, to reduce redundancy in the raw 50 Hz video streams, we applied uniform downsampling to 2.5 Hz. The sampled data were then segmented into video sequences according to the terrain categories listed in Table~\ref{tab:dataset_divide}. For the next stage of annotation, we followed a protocol designed to produce pixel-level, high-fidelity ground truth, categorizing each pixel into \textit{freespace} and \textit{background} regions. Our annotation pipeline consisted of three stages: (1) semi-automatic annotation of 4,000 images using X-AnyLabeling\cite{X-AnyLabeling} to train a small model; (2) model-based automatic segmentation of the remaining frames; and (3) verification and re-annotation of incorrect predictions using the semi-automatic tool.

Finally, as shown in Table~\ref{tab:dataset_divide}, we generate the IRON dataset containing 24,314 annotated frames from 35 video sequences. To prevent data leakage, we split the dataset at the sequence level, assigning 27 sequences to the training set and 8 sequences to the test set. This strict sequence-based separation ensures that models are evaluated on entirely unseen scenes.
\begin{table}[!t]
    \caption{Statistics of Our IRON Dataset.}
    \label{tab:dataset_divide}
    \centering
    \begin{threeparttable}
    \renewcommand{\arraystretch}{1.4}
    \resizebox{0.48\textwidth}{!}{
    \begin{tabular}{l|ccc|cc|c}    
        \hline
        \textbf{Category} & \textbf{C} & \textbf{F} & \textbf{H} & \textbf{B.L} & \textbf{L.L} & \textbf{Total} \\ 
        \hline
        \textbf{Train} & 5,169 & 7,746 & 7,107 & 10,321 & 9,701 & 20,022 \\
        \textbf{Test} & 651 & 1,886 & 1,755 & 3,104 & 1,188 & 4,292 \\
        \textbf{Total} & 5,820 & 9,632 & 8,862 & 13,425 & 10,889 & 24,314 \\
        \hline
        \textbf{Train Seq} & 1--10 & 13--25 & 30--33 & None & None & 27\\
        \textbf{Test Seq} & 11--12 & 26--29 & 34--35 & None & None & 8\\
        \hline
    \end{tabular}
    }
    \vspace{2mm}
    \begin{tablenotes}
        \footnotesize
        \item \hspace*{-2.7mm} \begin{minipage}{0.46\textwidth}
        \textbf{NOTE:} 
        C: Countryside; F: Forest; H: High Altitude; 
        B.L: Bright Light; L.L: Low Light (dusk, night). 
        Train/Test sequences for B.L and L.L are None because the dataset is organized by scene. 
        \end{minipage}
    \end{tablenotes}
    \end{threeparttable}
    \vspace{-5mm}
\end{table}

\vspace{-1mm}
\section{Proposed Approach}

\subsection{Overall Architecture} 
\begin{figure*}[!t]
    \centering
    \includegraphics[width=7.1in]{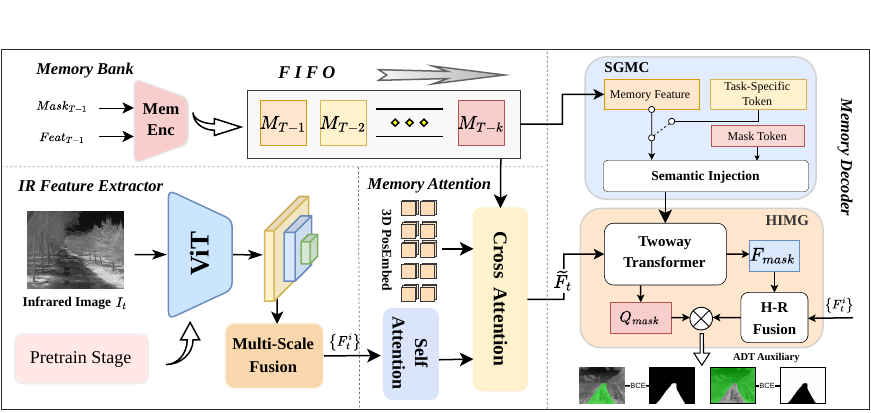}
    \caption{Overview of our proposed IRONet architecture. SGMC and HIMG represent the semantic-guided memory compensation and the hierarchical interactive mask generation modules. Task-Specific Token refers to the trainable freespace semantic token and background semantic token. H-R Fusion integrates upsampled features with the high-resolution features from $\{F_t^i\}$. ADT represents the alternating dual-task training strategy, which is active only during training and disabled at inference.}
    \label{fig:model_overview}
\end{figure*}

As illustrated in Fig.~\ref{fig:model_overview}, IRONet consists of three stages. A ViT backbone with a feature pyramid neck extracts multi-scale features from the input frame $I_t$. A memory attention module then queries a historical memory bank to produce a mask-aware temporal feature representation. Finally, a memory decoder generates the freespace mask from the enriched features. Each component is detailed in the following subsections.

\subsection{Multi-Scale Infrared Features} We build the feature extractor on a ViT backbone pretrained via masked autoencoding, which provides semantically rich yet spatially localized representations suited to infrared off-road imagery. To balance semantic consistency with boundary precision, we apply top-down multi-scale fusion: the deepest features are first processed by a pyramid spatial pooling module to consolidate global context, suppressing fragmentation caused by minor obstacles such as grass, then progressively fused with shallower spatial features to recover fine-grained boundary detail. The resulting feature pyramid $\{F_t^i\}$, indexed by level $i$, serves as the shared representation for both temporal aggregation and mask decoding.

\vspace{-2mm}
\subsection{Memory Attention and Bank}
We introduce a flow-free memory bank with cross-attention for lightweight temporal aggregation. The bank stores the highest-level historical features $F_{t-1}^0$ alongside their corresponding masks, encoded into mask-aware memory tokens via depthwise-separable convolution. These tokens jointly preserve visual and semantic cues and are maintained in a FIFO queue. To ensure diverse temporal coverage during training, the queue is populated via multi-interval sampling and augmentation, while inference uses the optimal interval. 

Memory attention blocks operate on top of the bank via an interleaved scheme. Current features are first refined by self-attention, then attend to memory tokens via cross-attention. 3D spatiotemporal positional embeddings, with the temporal dimension derived from timestamps, are injected prior to cross-attention to explicitly encode spatiotemporal relationships. Stacked blocks progressively yield a temporally enriched, mask-aware representation $\tilde{F}_t$. 

To bridge the training-inference gap, we adopt a curriculum strategy over historical masks. Training begins with teacher forcing on ground-truth masks and gradually transitions to an autoregressive regime where predicted masks are stored in memory, mitigating exposure bias and stabilizing inference. The memory bank is reset at sequence boundaries to ensure independence.

\vspace{-2mm}
\subsection{Memory Decoder}
The memory decoder takes the temporally-enriched, mask-aware features $\tilde{F}_t$ as input and produces the final freespace mask. Following the SAM2~\cite{ravi2024sam} paradigm, a learnable \textit{mask token} first interacts with historical freespace features from the memory bank to acquire target-class semantics, then queries the temporally enriched features $\tilde{F}_t$ through a bidirectional transformer, and finally produces the freespace mask via inner-product with the interaction-enhanced image features. While conceptually similar in structure, directly applying this paradigm to single-class infrared video segmentation introduces two non-trivial failure modes, motivating the three components described below.
\subsubsection{Semantic Guided Memory Compensation (SGMC)} The \textit{mask token} relies on mask-aware temporal features $\tilde{F}_t$ to identify the target class. When the memory bank is populated exclusively by freespace-empty frames, however, $\tilde{F}_t$ carries no reliable freespace semantics, leaving the \textit{mask token} without a valid segmentation target. This failure mode arises naturally at sequence initialization, during occlusions, and through sharp turns common in off-road driving. We address this with a quality-conditioned semantic injection mechanism. A mask-aware monitor evaluates the freespace coverage ratio across stored historical frames. When the coverage falls below a threshold $\tau$ (empirically set to 5\% of the image area), the decoder interprets the memory as unreliable and activates semantic compensation. A trainable \textit{freespace token} or \textit{background token} is concatenated with the \textit{mask token} to re-introduce class-specific guidance. This re-initialization enables the decoder to re-acquire the freespace target and maintain stable tracking through visually disruptive segments. Notably, at sequence onset the memory bank is initialized with empty masks, so SGMC naturally bootstraps segmentation without manual prompts, in contrast to SAM2~\cite{ravi2024sam} which requires explicit point-based initialization. 
\subsubsection{Hierarchical Interactive Mask Generation (HIMG)} Given the compensated \textit{mask token}, a multi-layer bidirectional transformer queries $\tilde{F}_t$ to aggregate mask evidence while propagating high-level context back into the image features. The resulting representation $F_{mask}$ is then refined through a hierarchical upsampling pathway, where transposed convolutions progressively recover spatial resolution and skip connections from the multi-scale feature pyramid $\{F_t^i\}$ reintroduce fine-grained boundary details. The final freespace mask is produced via inner-product between the enhanced query $Q_{mask}$ and the fused high-resolution features. \subsubsection{Alternating Dual-task Training (ADT)} Adapting the SAM2-like decoder to a single-class task introduces a \textit{semantic shortcutting} problem. In the original multi-class formulation, the \textit{mask token} is a class-agnostic query whose semantics are derived entirely from mask-aware $\tilde{F}_t$, providing strong gradient signal to the memory modules. Under single-class training, the \textit{mask token} collapses to a fixed freespace embedding, bypassing the memory modules and severing their supervision signal. To preserve class-agnosticism, we alternate the segmentation target across training clips between freespace and its complement. Formally, for each clip we define the target mask as \begin{equation} Y'_t = \begin{cases} Y_t, & \text{foreground task,} \\ 1 - Y_t, & \text{background task,} \end{cases} \end{equation} where $Y_t \in \{0,1\}^{H \times W}$ is the ground-truth freespace mask. Task assignment is sampled proportionally to class frequency in the training set. The model is then trained end-to-end with binary cross-entropy: \begin{equation} \label{eq:bce_loss} \mathcal{L} = -\,\mathbb{E}\!\left[Y'_t \log P_t + (1-Y'_t)\log(1-P_t)\right], \end{equation} where $P_t \in [0,1]^{H \times W}$ is the predicted probability map. By forcing the \textit{mask token} to represent both semantic classes across training, ADT prevents shortcutting, maintains strong supervision to the temporal memory modules, and improves overall temporal consistency.

\begin{table*}[!t]
    \caption{Comparison of camera-only methods on IRON Dataset under the infrared modality.}
    \label{tab:otdr_comparison}
    \centering
    \renewcommand{\arraystretch}{1.5}
    \begin{threeparttable}
    \begin{tabular}{>{\bfseries}l|ccccccc}
    \hline
    \textbf{Methods} & \textbf{Backbone} & \textbf{Precision}(\%) & \textbf{Recall}(\%) & \textbf{F1}(\%)  & \textbf{IoU}(\%) &  \textbf{Params}(M) & \textbf{FPS} \\
    \hline
    U-Net~\cite{ronneberger2015u} & - & 71.30 & 90.12 & 79.61 & 66.13 & 31.04 & 21 \\

    SegFormer~\cite{xie2021segformer} & ViT-S & 82.75 & \underline{92.41} & 87.31 & 77.48 & 27.48 & 33 \\

    ROD~\cite{sun2025rod} & ViT-S & 86.12 & 89.06 & 87.56 & 77.88 & 33.43 & 68 \\

    DeepLabV3+~\cite{chen2018encoder} & ResNet-101 & 87.25 & 90.31 & 88.75 & 79.78 & 58.75 & 103 \\

    Mask2Former~\cite{cheng2022masked} & ResNet-50 & 87.80 & 92.22 & 89.95 & 81.74 & 43.95 & 23 \\
    \hline
    IRONet\_3F & ViT-B & \underline{88.15} & \textbf{93.07} & \underline{90.55} & \underline{82.73} & 104.49 & 23 \\
    IRONet\_5F & ViT-S &  \textbf{90.85} & 90.49 & \textbf{90.66} & \textbf{82.93} & 40.05 & 32 \\
    \hline
    \end{tabular}
    \vspace{2mm}
    \begin{tablenotes}
        \footnotesize
        \item \hspace*{-3.2mm} \begin{minipage}{0.71\textwidth} \textbf{NOTE:} IRONet\_3F represents our backbone with a three-frame memory queue, while IRONet\_5F extends it to five frames, capturing longer-range temporal dependencies. FPS is measured on a single A100 GPU. 
        \end{minipage}
    \end{tablenotes}
    \end{threeparttable}
    \vspace{-4mm}
\end{table*}




\section{Experiments}

\subsection{Implementation Details}

All experiments are conducted on our IRON dataset or public benchmarks using the official training and testing splits. Our implementation is based on the mmsegmentation framework. During training, input frames are augmented via random scaling, cropping, and horizontal flipping, and resized to $512 \times 512$, with identical transformations applied across frames to preserve temporal consistency. The model is trained for 10 epochs on a single A100 GPU using AdamW with a learning rate of $1\text{e-}4$, weight decay of $0.05$, and a batch size of 16. The default memory queue length is set to $L=3$, while the best-performing IRONet\_5F adopts $L=5$ as determined in our ablation study. During inference, images are processed in a streaming manner at their original resolution of $640 \times 512$. Performance is evaluated using IoU, Precision, Recall, and F1-score.

\subsection{Main Results}
 Table~\ref{tab:otdr_comparison} reports results on the IRON test set against CNN-based, Transformer-based, and off-road specialist baselines. IRONet leads across all metrics, with the performance gap most pronounced over strong single-frame competitors such as Mask2Former and ROD. We attribute this to a fundamental limitation of frame-independent methods. In low-contrast infrared imagery, local appearance is often insufficient to resolve freespace boundaries, leading to spatial fragmentation and temporal flickering. IRONet addresses this at the architectural level by aggregating multi-frame context through memory attention, converting ambiguous per-frame evidence into coherent, stable predictions. Efficiency scales favorably alongside accuracy. IRONet\_5F operates at 32 FPS with 40.05M parameters, lighter than Mask2Former and faster than most competitors, confirming that flow-free temporal reasoning imposes no meaningful computational overhead over single-frame inference.

\subsection{Generalization to RGB Scenarios} To assess modality generalizability of our method, we evaluate IRONet on ORFD and Rellis-3D without any domain-specific adaptation. As shown in Tables~\ref{tab:orfd} and~\ref{tab:rellis}, IRONet consistently outperforms all RGB-only baselines and, notably, surpasses RGB+LiDAR competitors on both benchmarks despite using only monocular camera input. These results confirm that the proposed memory-attention mechanism captures modality-agnostic temporal structure, and that IRONet effectiveness is not contingent on infrared-specific appearance cues.

\begin{table}[h]
    \centering
    \renewcommand{\arraystretch}{1.3}
    \caption{Comparison on the ORFD dataset.}
    \label{tab:orfd}
    \begin{threeparttable}
    \begin{tabular}{c l cccc}
    \hline
     & \textbf{Methods} & \textbf{Precision} & \textbf{Recall} & \textbf{F1} & \textbf{IoU} \\
    \hline
    \multirow{5}{*}{\raisebox{-.5\height}{\rotatebox{90}{\textit{RGB+LiDAR}}}} 
    & \textbf{RTFNet}~\cite{sun2019rtfnet} & 84.2 & 96.7 & 90.0 & 81.8 \\
    & \textbf{SNE-RoadSeg}~\cite{fan2020sne} & 86.7 & 92.7 & 89.6 & 81.2 \\
    & \textbf{OFF-Net}~\cite{min2022orfd} & 86.6 & 94.3 & 90.3 & 82.3 \\
    & \textbf{M2F2-Net}~\cite{ye2023m2f2} & 97.3 & 95.5 & 96.4 & 93.1 \\
    & \textbf{RoadFormer}~\cite{li2024roadformer} & 95.1 & 97.2 & 96.1 & 92.5 \\
    \hline
    \multirow{5}{*}{\raisebox{-.5\height}{\rotatebox{90}{\textit{RGB-only}}}} 
    & \textbf{U-Net}~\cite{ronneberger2015u} & 63.7 & 53.7 & 58.3 & 41.1 \\
    & \textbf{DeepLabV3+}~\cite{chen2018encoder} & 78.1 & 87.1 & 82.4 & 70.0 \\
    & \textbf{ROD}~\cite{sun2025rod}\textsuperscript{†} & 97.9 & 96.3 & 97.1 & 94.3 \\
    & \textbf{IRONet\_3F} & \underline{98.0} & \underline{96.5} & \underline{97.2} & \underline{94.6} \\
    & \textbf{IRONet\_5F} & \textbf{98.0} & \textbf{96.7} & \textbf{97.3} & \textbf{94.8} \\
    \hline
    \end{tabular}
    \vspace{2mm}
    \begin{tablenotes}
        \footnotesize
        \item \textbf{NOTE:} All results are based on a ViT-S backbone. \textsuperscript{†} reproduced from official code.
    \end{tablenotes}
    \end{threeparttable}
    \vspace{-2mm}
\end{table}
\begin{table}[h]
    \centering
    \renewcommand{\arraystretch}{1.3}
    \caption{Comparison on the Rellis-3D dataset.}
    \label{tab:rellis}
     \begin{threeparttable}
    \begin{tabular}{l | c c c c}
    \hline
    \textbf{Method} & \textbf{Modality} & \textbf{Precision} & \textbf{Recall} & \textbf{F1} \\
    \hline
    \textbf{U-Net~\cite{ronneberger2015u}}                  & RGB        & 80.90  & 85.60  & 83.20  \\
    \textbf{DeepLabV3+~\cite{chen2018encoder}}              & RGB        & 58.60  & 42.80  & 49.50  \\
    \textbf{ROD~\cite{sun2025rod}\textsuperscript{†}}                          & RGB        & 94.70 & \underline{95.70} & 95.20 \\
    \hline
    \textbf{Real-NVP~\cite{wellhausen2020safe}}             & RGB+LiDAR  & 57.10 & 97.42 & 70.01 \\
    \textbf{AE Based~\cite{schmid2022self}}                 & RGB+LiDAR  & 70.79 & 91.81 & 74.37 \\
    \textbf{SEO~\cite{seo2023learning}}                     & RGB+LiDAR  & 91.64 & 85.08 & 86.22 \\
    \textbf{M2F2-Net~\cite{ye2023m2f2}}                     & RGB+LiDAR  & 92.50  & 96.40  & 94.40  \\
    \hline
    \textbf{IRONet\_3F} & RGB & 94.51 & \textbf{95.87} & 95.18 \\
    \textbf{IRONet\_5F} & RGB & \textbf{95.65} & 95.69 & \textbf{95.67} \\
    \hline
    \end{tabular}
    \vspace{2mm}
    \begin{tablenotes}
        \footnotesize
            \item \hspace*{-4mm} \begin{minipage}{0.45\textwidth} \textbf{NOTE:} All results of our method are obtained using a ViT-S backbone. 
        \end{minipage}
    \end{tablenotes}
    \end{threeparttable}
    \vspace{-2mm}
\end{table}

\subsection{Ablation Studies}
To validate our design of our framework, we conduct a series of ablation studies to quantify the contribution of the IRONet framework's principal components. Unless otherwise specified, all experiments are performed using the ViT-S backbone initialized with ConvMAE~\cite{gao2022convmae} pre-trained weights.

\subsubsection{Effectiveness of Temporal Modules} Table~\ref{tab:temporal_abl} incrementally validates each component of the temporal fusion module. The baseline without any temporal modeling achieves 79.10\% IoU, already competitive among single-frame methods. Introducing memory attention yields the largest individual gain, confirming that historical feature aggregation is the primary driver of improvement in low-contrast infrared scenes where per-frame appearance alone is ambiguous. ADT further regularizes temporal representation learning by keeping the mask token class-agnostic, preventing semantic shortcutting and strengthening supervision signals to the memory modules. SGMC addresses the remaining failure mode, tracking loss under freespace-empty frames, by re-initializing decoder semantics on demand. The full model achieves 82.28\% IoU, with each component contributing a consistent and complementary gain.

\begin{table}[!h]
    \caption{Ablation of Temporal Modules.}
    \label{tab:temporal_abl}
    \centering
    \renewcommand{\arraystretch}{1.4}
    \begin{threeparttable}
    \begin{tabular}{ccc|cccc}
    \hline
    \textbf{MA} & \textbf{ADT} & \textbf{SGMC} & \textbf{Precision} & \textbf{Recall} & \textbf{F1} & \textbf{IoU} \\
    \hline
    - & - & - & 84.32 & \textbf{92.74} & 88.33 & 79.10 \\

    \checkmark & - & - & 87.37 & 91.74 & 89.50 & 81.00    \\

    \checkmark & \checkmark & - & 88.07 & 91.54 & 89.77 & 81.44 \\
    \checkmark & \checkmark & \checkmark &  \textbf{88.77} & \underline{91.84} & \textbf{90.28}& \textbf{82.28} \\
    \hline
    \end{tabular}
    \vspace{2mm}
    \begin{tablenotes}
        \footnotesize
        \item \hspace*{-2.9mm} \begin{minipage}{0.42\textwidth} \textbf{NOTE:} MA: Memory Attention; ADT: Alternating Dual-task Training; SGMC: Semantic Guided Memory Compensation.
        \end{minipage}
    \end{tablenotes}
    \end{threeparttable}
\end{table}

\begin{figure*}[!t]
    \centering
    \includegraphics[width=\textwidth]{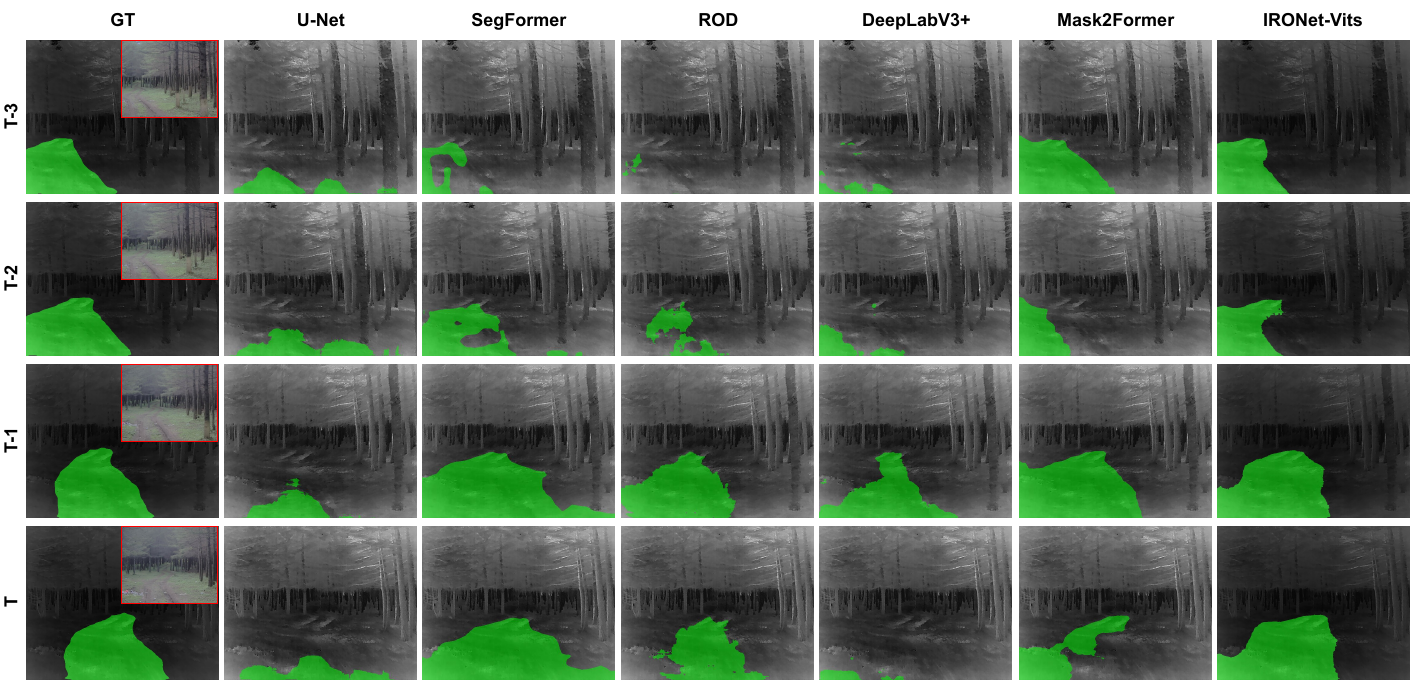}
    \caption{Qualitative comparison of IRONet against state-of-the-art methods on a representative sequence from the IRON test set. Rows allow cross-method inspection at fixed timesteps, while columns reveal the temporal coherence of each method across consecutive frames.}
    \label{fig:vis_road}
    \vspace{-2mm}
\end{figure*}

\begin{table}[!h]
    \caption{Ablation of Memory Bank Length.}
    \label{tab:queue_abl}
    \centering
    \renewcommand{\arraystretch}{1.4}
    \begin{tabular}{c|cccc}
    \hline
    \textbf{Queue Length} & \textbf{Precision} & \textbf{Recall} & \textbf{F1} & \textbf{IoU} \\
    \hline
    3 Frames & 88.77 & 91.84 & 90.28 & 82.28 \\
    5 Frames & \textbf{90.85} & \underline{90.49} & \textbf{90.66} & \textbf{82.93} \\
    7 Frames & 87.93 & 91.82 & 89.83 & 81.54 \\
    \hline
    \end{tabular}
    \vspace{-2mm}
\end{table}

\subsubsection{Impact of Memory Bank Length} As shown in Table~\ref{tab:queue_abl}, performance peaks at $L\!=\!5$ and degrades symmetrically in both directions. A shorter window undersamples temporal context, while an excessively long one introduces temporally distant features that act as noise rather than signal. This non-monotonic trend indicates the existence of an optimal temporal horizon, beyond which historical information becomes detrimental to segmentation consistency.

\begin{table}[!h]
    \caption{Ablation of Backbone.}
    \label{tab:backbone_abl}
    \centering
    \renewcommand{\arraystretch}{1.4}   
    \begin{threeparttable}
        \begin{tabular}{l|ccccc}
        \hline
        \textbf{Pretrained} & \textbf{Backbone} & \textbf{Precision} & \textbf{Recall} & \textbf{F1} & \textbf{IoU} \\
        \hline
        None & ViT-S & 86.98 & 86.96 & 86.97 & 76.94 \\
        Dinov3~\cite{simeoni2025dinov3} & ViT-S & 89.88 & 90.52 &  90.20 & 82.15  \\
        ConvMAE~\cite{gao2022convmae} & ViT-S & 88.77 & 91.84 & 90.28 & 82.28  \\
        \hline
        None & ViT-B& 85.69 & 88.37 & 87.01 & 77.01 \\ 
        Dinov3~\cite{simeoni2025dinov3} & ViT-B & \textbf{89.34} & \underline{91.62} & \underline{90.46} & \underline{82.59} \\
        ConvMAE~\cite{gao2022convmae} & ViT-B & \underline{88.15} & \textbf{93.07} & \textbf{90.55} & \textbf{82.73} \\
        \hline
        \end{tabular}
        \vspace{2mm}
        \begin{tablenotes}
            \footnotesize
            \item \hspace*{-3mm} \begin{minipage}{0.46\textwidth} \textbf{NOTE:} Pretrained represents the method of pretrained model we used. None indicates that the model is trained from scratch without using any pretrained weights. Specifically, DINOv3 is pretrained on LVD-1689M~\cite{simeoni2025dinov3} and ConvMAE is pretrained on ImageNet-1K~\cite{deng2009imagenet}.
            \end{minipage}
        \end{tablenotes}
    \end{threeparttable}
\end{table}

\subsubsection{Impact of Backbone and Pre-training} As shown in Table~\ref{tab:backbone_abl}, pre-training is indispensable: training from scratch degrades IoU by over 5\%, confirming that strong initial representations are a prerequisite for effective temporal fusion. Among pre-training strategies, ConvMAE consistently outperforms DINOv3, suggesting that the masked autoencoding objective better aligns with the local structural reasoning required by memory attention. 
\subsection{Qualitative Analysis} Fig.~\ref{fig:vis_road} reveals a systematic behavioral difference between single-frame methods and IRONet. Across frames $T\!-\!3$ to $T$, single-frame predictions exhibit boundary flickering and spatial fragmentation, artifacts that quantitative metrics partially obscure. IRONet produces temporally stable, spatially coherent masks throughout, as memory-based context aggregation continuously suppresses transient noise and resolves per-frame ambiguities. Such temporal consistency is critical for safety-sensitive autonomous navigation, where prediction instability directly impacts downstream planning.

\section{Conclusion}
This work presented IRON, the first large-scale infrared video dataset for temporal freespace detection in off-road environments, and IRONet, a temporal segmentation framework based on flow-free memory attention. By combining a mask-aware temporal queue with alternating dual-task training and semantic guided memory compensation, IRONet aggregates multi-frame context to produce freespace masks with improved temporal stability and spatial completeness across diverse terrains at real-time speed. Experiments and ablation studies on IRON show that our method achieves state-of-the-art performance, confirm the benefit of explicit temporal modeling, and provide a reliable basis for future research on infrared off-road perception and temporal fusion. Future work will focus on extending the dataset toward all-weather, long-term off-road freespace detection with multi-modal fusion.

\bibliographystyle{IEEEtran}
\bibliography{IEEEabrv,IEEEexample}

\end{document}